\documentclass{article} 
\usepackage{iclr2026_conference,times}
\iclrfinalcopy
\usepackage{listings}
\usepackage{xcolor}
\usepackage{float}

\usepackage{amssymb}
\usepackage{makecell}
\usepackage{amssymb,graphicx}
\newcommand{\RightAngle}{%
  \rotatebox[origin=c]{270}{\scalebox{1.2}{$\lrcorner$}}%
}
\usepackage{tikz}
\usetikzlibrary{arrows.meta, positioning, fit, backgrounds}

\usepackage{listings}

\definecolor{keywordcolor}{rgb}{0.7, 0.1, 0.1}   
\definecolor{tacticcolor}{rgb}{0.0, 0.1, 0.6}    
\definecolor{commentcolor}{rgb}{0.3, 0.5, 0.3}   
\definecolor{symbolcolor}{rgb}{0.0, 0.1, 0.6}    
\definecolor{sortcolor}{rgb}{0.1, 0.5, 0.1}      
\definecolor{attributecolor}{rgb}{0.7, 0.1, 0.1} 
\definecolor{rulecolor}{rgb}{0, 0, 0}

\usepackage{amsmath,amsfonts,bm}

\def\eqref#1{equation~\ref{#1}}

\def\1{\bm{1}}

\DeclareMathAlphabet{\mathsfit}{\encodingdefault}{\sfdefault}{m}{sl}
\SetMathAlphabet{\mathsfit}{bold}{\encodingdefault}{\sfdefault}{bx}{n}

\usepackage{hyperref}
\usepackage{url}

\title{LeanGeo: Formalizing Competitional Geometry problems in Lean}

\author{
Chendong Song\thanks{Equal contribution.}~\,\textsuperscript{1} ~~
Zihan Wang\footnotemark[1]~\,\textsuperscript{13} \quad
Frederick Pu\textsuperscript{24} \quad
Haiming Wang\textsuperscript{1} \quad
Xiaohan Lin\textsuperscript{1} \\
~\textbf{Junqi Liu\textsuperscript{1} \quad
Jia Li\textsuperscript{2} \quad
Zhengying Liu\thanks{Corresponding author.}}~\,\textsuperscript{1} \\
\textsuperscript{1}Moonshot~AI \quad \textsuperscript{2}Numina \quad \textsuperscript{3}Peking University \quad \textsuperscript{4}University of Toronto\\
\texttt{songcd1212@gmail.com, zihanwang@stu.pku.edu.cn,} \\\texttt{liuzhengying@moonshot.cn}}

\begin{document}

\maketitle

\begin{abstract}
Geometry problems are a crucial testbed for AI reasoning capabilities. Most existing geometry solving systems cannot express problems within a unified framework, thus are difficult to integrate with other mathematical fields. Besides, since most geometric proofs rely on intuitive diagrams, verifying geometry problems is particularly challenging. To address these gaps, we introduce LeanGeo, a unified formal system for formalizing and solving competition-level geometry problems within the Lean 4 theorem prover. LeanGeo features a comprehensive library of high-level geometric theorems with Lean’s foundational logic, enabling rigorous proof verification and seamless integration with Mathlib. We also present LeanGeo-Bench, a formal geometry benchmark in LeanGeo, comprising problems from the International Mathematical Olympiad (IMO) and other advanced sources. Our evaluation demonstrates the capabilities and limitations of state-of-the-art Large Language Models on this benchmark, highlighting the need for further advancements in automated geometric reasoning. We open source the theorem library and the benchmark of LeanGeo at \url{https://github.com/project-numina/LeanGeo/tree/master}
\end{abstract}



\section{Introduction}



In recent years, Large Language Models (LLMs) have made significant progress in mathematical reasoning, particularly in automated theorem proving~\citep{bibel2013automated}. Formal theorem proving is a crucial domain for ensuring the correctness of hard-to-verify proofs within theorem proving. Lean 4~\citep{moura2021lean}, as a prominent proof assistant, provides a solid foundation for algebra and number theory through its extensive Mathlib library~\citep{mathlib2020}. It has been widely used in the formal verification of theorems within LLMs. 

However, Euclidean geometry, an essential component of mathematical reasoning and a frequent focus of competitions, remains relatively underexplored in Lean 4 community, Mathlib and automated theorem provers. This stems from the inherent difficulty of geometric problems, which demand graphic intuition; human reasoning in such cases inevitably relies on geometric insight, making absolute formalization of geometry problem extremely challenging.

Currently, advanced geometric systems like AlphaGeometry~\citep{trinh2024solving}, TongGeometry~\citep{zhang2024proposing} and SeedGeometry~\citep{chen2025seedproverdeepbroadreasoning},  while achieving impressive results on IMO-level geometry problems, typically rely on specialized models and operate within geometry-specific formal systems independent of Lean. This isolation prevents integration with other mathematical domains in mathlib, making it hard to tackle geometry problems that intersect with algebra, number theory, or combinatorics, which are common in recent competitions. It is also difficult to apply tools from other areas, such as trigonometry or inequalities, in geometric proofs. Additionally, their reliance on graphical verification and unordered formal systems can lead to logical unsoundness and incompleteness. 

Even in Lean 4, geometric results remain scarce: Mathlib’s formalized geometry is algebraic, offering little support for the kind of logical, diagram-based reasoning typical of geometry. Recently, LeanEuclid~\citep{murphy2024autoformalizing} established a geometric axiomatic system based on SystemE~\citep{avigad2009formal} and established an autoformalization benchmark. However, their theorem library only covers the first chapter of Elements, leaving it unable to express most of the middle-school level geometry content. 

While developing a robust formal system is a vital step toward rigorous geometric reasoning, equally important is the establishment of suitable benchmarks to rigorously evaluate the geometric reasoning capability of LLMs. However, since most geometric proofs rely on intuitive diagrams, verifying geometry problems is particularly challenging. Existing geometry benchmarks, such as Geoeval~\citep{zhang2024geoeval}, GeoQA~\citep{chen2021geoqa} Geometry3K~\citep{geometry3k_dataset} and FormalMath~\citep{yu2025formalmath}, primarily emphasize numerical computations of geometry object, focusing on models’ computational ability rather than their true geometric reasoning skills. MATP-BENCH \cite{he2025matp} comprises 1,056 geometry problems of varying difficulty levels implemented in Lean4. However, current LLMs exhibit unsatisfactory performance on this benchmark. Due to the absence of a comprehensive geometry theorem library in Lean4, it is challenging for LLMs to effectively apply geometric tools within this environment. This highlights the necessity of developing a complete formal system and an extensive theorem library to serve as reliable tools for LLMs.

To handle these critical gaps, we introduce LeanGeo, a framework designed to formalize and solve geometric problems in Lean 4. Building upon LeanEuclid~\citep{murphy2024autoformalizing}, LeanGeo establishes a comprehensive library of geometric theorems specifically curated for competition-level challenges and seamlessly integrates with Mathlib.
Compared to other formal systems like AlphaGeometry, LeanGeo exhibits significant differences, as detailed in Table \ref{ComparisonAG} and Appendix \ref{Appendix:Comparison}.

\vspace{-0.5cm}
\begin{table}[h]
\caption{Comparison of problem with AlphaGeometry and LeanGeo}
\vspace{-0.4cm}
\label{ComparisonAG}
\begin{center}
\begin{tabular}{|l|l|}

\hline

Natural Language & \makecell[lc]{
In a triangle $ABC$, side $AB = AC$, prove that $\angle ACB = \angle ABC$. \\ \\
\textbf{Solution.} Choose $D$ as the midpoint of side $BC$. Then $\triangle ABD$ and $\triangle ACD$ \\are congruent. Therefore, $\angle ACB = \angle ACD = \angle ABD = \angle ABC$ 
\\
}\\
\hline
AlphaGeometry & \makecell[lc]{
a b = segment a b; c = on\_circle c a b ? eqangle b a b c b c c a \\ \\
\textbf{Solution.} * From theorem premises:\\
A B C : Points\\
cong A C A B [00]\\
 * Auxiliary Constructions:\\
: Points\\
 * Proof steps:\\
001. cong A C A B [00] $\Rightarrow$ eqangle A C B C B C A B
\\
}\\
\hline
LeanGeo & 
\makecell[lc]{
theorem isoTriangle\_imp\_eq\_angles : $\forall$ (A B C : Point),  \\
IsoTriangle A B C $\rightarrow$ $\angle$ A:B:C =$\angle$ A:C:B := by \\
\quad euclid\_intros \\
\quad  euclid\_apply exists\_midpoint B C as D \\
\quad  euclid\_apply line\_from\_points B C as BC \\
\quad  euclid\_apply coll\_angles\_eq \\
\quad  euclid\_apply congruentTriangles\_SSS D B A D C A \\
\quad  euclid\_apply coll\_angles\_eq \\
\quad  euclid\_finish \\
}
\\
\hline
\end{tabular}
\end{center}
\end{table}
\vspace{-0.5cm}
Based on this theorem library, we propose LeanGeo-Bench, the first formalized geometric problem benchmark in Lean 4. It comprises 122 geometry problems, including all International Mathematical Olympiad (IMO) geometry problems since 2000. Furthermore, we present a training methodology that uses the theorem library to construct supervised fine-tuning (SFT) data. This data is then used in reinforcement learning (RL) experiments upon the Kimi k1.5 reinforcement learning (RL) pipeline \citep{team2025kimi}, yielding promising initial results.

The primary contributions of this work are as follows:
\begin{itemize}
    \item We present the first framework in the Lean theorem prover capable of expressing and reasoning about competition-level geometry problems in a human-like manner. The framework features an extensive library of high-level definitions and tactics based on theorems commonly used by IMO competitors, making formal proofs more intuitive and understandable. Its integration within Lean facilitates the formalization of problems at the intersection of geometry and other domains like combinatorics.
    \item We introduce a comprehensive geometry benchmark formalized in Lean 4 and LeanGeo, capable of representing most of the geometry problems from the International Mathematical Olympiad (IMO). This benchmark provides a standardized and challenging testbed for evaluating future formal mathematics systems. We also provide baseline results on this benchmark using several state-of-the-art large language models.
\end{itemize}

\section{Related Work}

\subsection{Automated Theorem Proving}
Interactive theorem provers span a spectrum of foundational languages:
HOL4~\citep{slind2008brief} and Isabelle/HOL~\citep{paulson1994isabelle} rely on simply-typed higher-order logic,
Coq~\citep{barras1999coq} and Lean~\citep{de2015lean} on dependent type theory.
Within this landscape, Lean~4 has emerged as the de-facto standard for new developments: its large and rapidly growing \lstinline{Mathlib4} library \citep{mathlib2020}, native-code execution, and seamless metaprogramming interface have attracted both mathematicians and AI researchers, making it the most actively used system for contemporary formalisation efforts.

In parallel, a series of search-based theorem provers have been developed to enhance automated reasoning capabilities. LEGO-Prover \citep{wang2023lego} employs a modular formal proof framework to construct a reusable skill library, enabling LLMs to retrieve existing skills and synthesise new ones during the proof process. DT-Solver \citep{wang2023dt} introduces a dynamic-tree Monte Carlo search algorithm, whereas BFS-Prover \citep{xin2025bfs}, based on a best-first search strategy, achieves state-of-the-art performance among search-based theorem provers.

More recent developments have shifted towards an alternative whole-proof generation approach, where a language model generates the entire proof in a single pass. Notable examples following this paradigm include DeepSeek-Prover \citep{ren2025deepseek}, Goedel-Prover \citep{lin2025goedel}, and Kimina-Prover Preview \citep{wang2025kimina}. These provers are predominantly trained on Lean  4-specific corpora and problem sets, enabling strong performance in algebraic and number-theoretic domains; however, their effectiveness diminishes when faced with tasks requiring complex reasoning or involving unfamiliar problem domains. In contrast, agentic methods such as Delta Prover\citep{delta_prover_2025} integrate reflective decomposition and iterative repair, allowing a general-purpose LLM to interactively construct formal proofs. Seed-Prover \citep{chen2025seedproverdeepbroadreasoning} combines multi-stage reinforcement learning, agent-based strategies and test-time scaling, achieving impressive results by fully solving 4 out of 6 problems in IMO 2025.

\subsection{LeanEuclid}
\label{sec:leaneuclid}
LeanEuclid~\citep{murphy2024autoformalizing} represents a pioneering effort in formalizing plane geometry within Lean by integrating SMT~\citep{barrett2018satisfiability} solving techniques with SystemE \citep{avigad2009formal} to construct a rigorous axiomatic framework. It introduces an autoformalization benchmark that covers the first chapter of Euclid’s \emph{Elements} along with 125 relatively simple problems drawn from the UniGeo corpus. Despite this progress, the formal system and benchmark remains limited in scope, primarily addressing foundational geometric propositions and exercises, and does not yet encompass the breadth of topics found in modern middle school geometry textbooks or the complexity typical of competition-level geometry problems.

\subsection{Geometry Problem Solving}
\label{sec:related:gps}
Automatic geometry solvers have a rich history.  Classical algebraic methods—Wu's characteristic set~\citep{wen1986basic} and Gröbner bases~\citep{bose1995grobner}—reduce geometry to polynomial ideal membership, achieving impressive coverage of textbook theorems.

A recent milestone in automated geometry reasoning is AlphaGeometry~\citep{trinh2024solving}, which integrates a neural language model trained on 100 million synthetic theorems with a symbolic deduction engine to solve 25 out of 30 IMO-level problems. Building on the framework proposed in~\cite{chou2000deductive}, its formal system is unordered and point-centered, enabling fast symbolic deduction within this setting. However, this formal system has several limitations. First, the deduction process heavily depends on specific graphs and coordinate choices, which makes it challenging to verify the correctness in this system. Second, the rule set employed by AlphaGeometry is low-level and difficult to extend, resulting in proof steps that are unnatural and hard for humans to interpret. Third, the formal system is isolated in scope, capable of addressing only a subset of geometry problems; consequently, it cannot incorporate external knowledge nor effectively handle interdisciplinary problems such as those involving geometric inequalities.

Our LeanGeo library bridges this gap by providing (i) an axiomatic, coordinate-free geometric library compatible with \lstinline{Mathlib}. (ii) a growing benchmark of competitional geometry problems with formal proofs.

\section{LeanGeo}
LeanGeo is a manually formalized system of plane geometry theorems and their proofs in the Lean 4 proof assistant. It builds upon the axiomatic framework of SystemE~\citep{avigad2009formal}, while its implementation inherits most foundational geometric objects, relations from LeanEuclid \citep{murphy2024autoformalizing}, with slight modifications (see Appendix \ref{Appendix:Change}). Additionally, LeanGeo leverages LeanSMT~\citep{mohamed2025lean} at its core, which effectively hides many of the underlying proof details in Lean 4.
\vspace{-0.1cm}
\begin{figure}[h]
    \centering
    \includegraphics[width=0.8\linewidth]{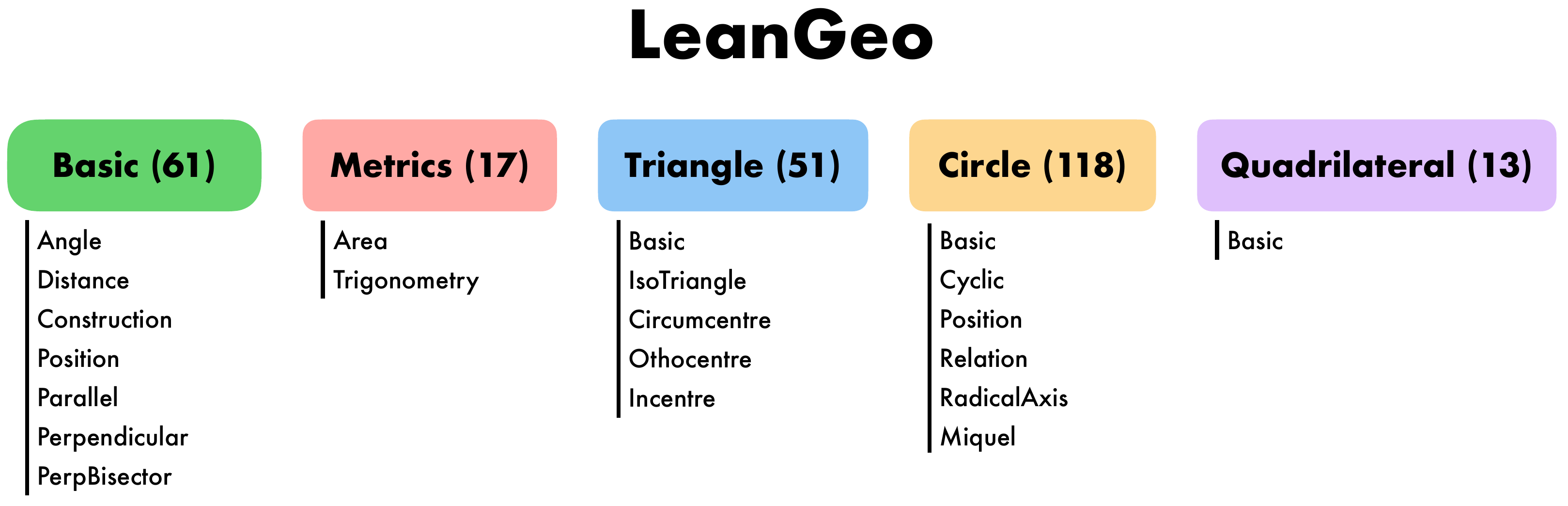}
    \caption{Structure of LeanGeo Theorem Library}
    \label{fig:Library}
\end{figure}
\vspace{-0.5cm}
\subsection{Theorem Library}

To enhance the expressive power of the theorem library and align it with common geometric terminology, we firstly introduced 52 new definitions for geometric structures — such as Midpoint, Circumcenter, and RadicalAxis using \verb|abbrev| as shown in \ref{lst:abbrev-example}. These additions make problem statements more concise and proofs more streamlined, while not increasing the length of the corresponding SMT process.
\begin{lstlisting}[caption={Example of abbreviation}, label={lst:abbrev-example},frame=single]
abbrev Cyclic (A B C D: Point) : Prop :=
 ∃ (O: Circle), A.onCircle O ∧ B.onCircle O ∧ C.onCircle O ∧ D.onCircle O
\end{lstlisting}

With the assistance of these newly defined structures, we established LeanGeo, a theorem library comprising 260 geometric theorems as shown in \ref{fig:Library}. All theorems in the library are manually written, formally proved and auto-verified by Lean4 and LeanSMT. 

These theorems systematically cover topics ranging from foundational middle-school geometry to challenging International Mathematical Olympiad (IMO) level theorem, such as Menelaus’s theorem and Miquel’s theorem. Besides, the library covers a wide range of geometry theorem, including fundamental properties of triangles (e.g., congruence, similarity), circles (e.g., inscribed angles, power of a point, radical axis), and quadrilaterals, as well as theorems related to key geometric points like the circumcenter and orthocenter. 

A key feature of LeanGeo is that most proofs in library are constructed by referencing previously established theorems through the \verb|euclid_apply| tactic. Consequently, the development of the library parallels the human approach to building geometric theory—progressing from simple foundations to increasingly complex structures. This feature is exemplified in Listing \ref{example_library}. As the collection of formally verified theorems and lemmas expands, the system’s deductive capacity correspondingly increases, allowing higher-level theorems to be proved more efficiently by leveraging existing results. In addition, the well-structured nature of LeanGeo’s proofs substantially lowers the entry barrier for mathematicians, geometry educators, and students. This structured and intuitive format also benefits LLMs, enabling them to parse, learn from, and even contribute to the formal library more effectively.

Besides, LeanGeo is designed for seamless integration with Mathlib, enabling it to leverage powerful tools from other areas of mathematics. For example, it can employ trigonometric identities and advanced inequalities to tackle problems that are often beyond the reach of purely axiomatic geometry systems. As shown in \ref{Appendix:imo2001}, trigonometric theorems in Mathlib are applied to prove IMO\_2001\_P1, a geometry inequality problem that is difficult to express within most geometric formal systems.

\begin{figure}[h]
    \centering
    \includegraphics[width=\linewidth]{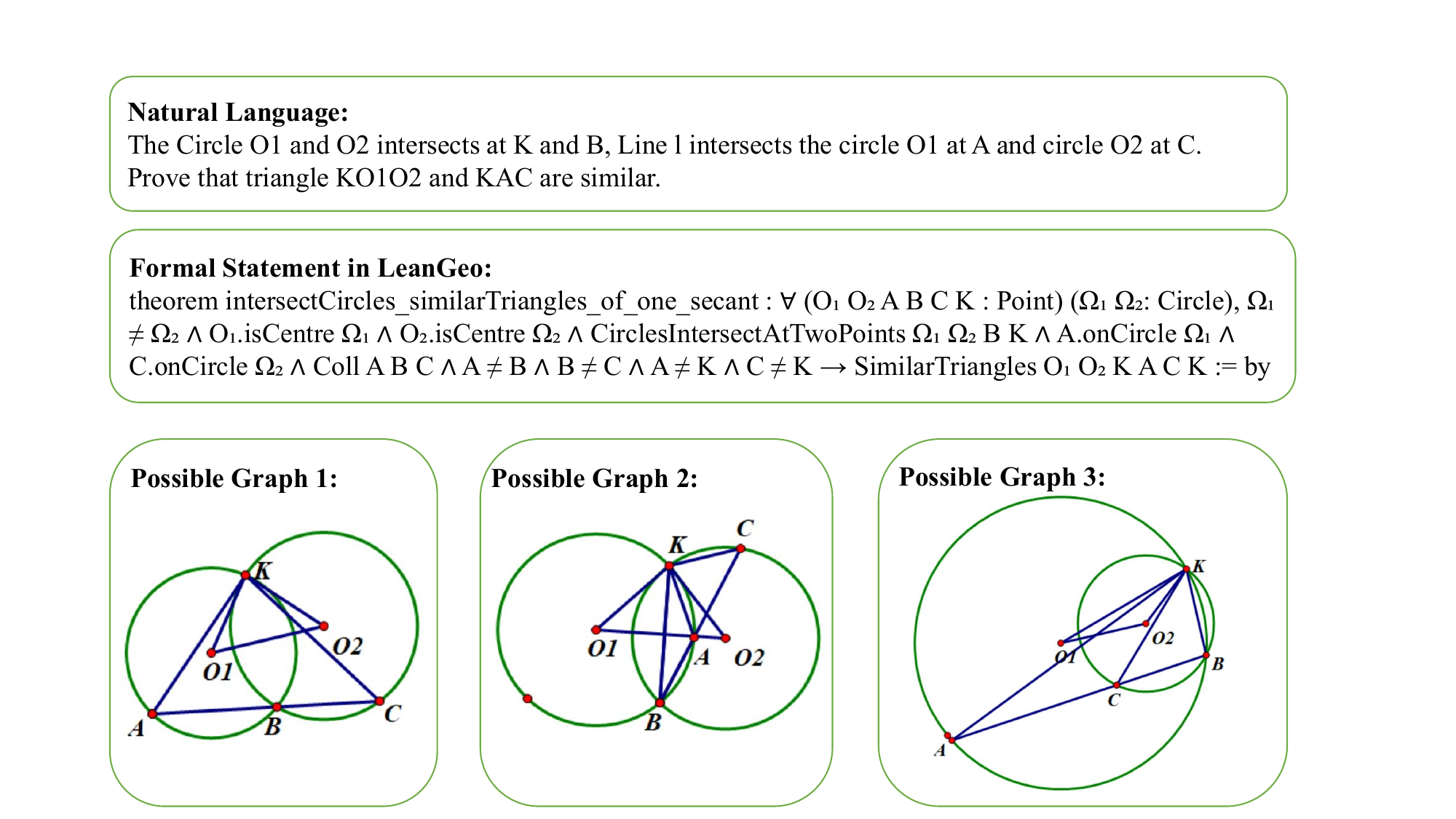}
    \vspace{-0.1cm}
    \caption{Different graphs with a same formal statement}
    \label{fig:Miq}
\end{figure}

One of the most challenging issues in theorem annotation is that describing positional relationships in geometry without visual aids is inherently complex. For problem illustrated in Figure \ref{fig:Miq}, natural language proofs, as well as most geometry formal systems such as AlphaGeometry, consider only a single case. Owing to Lean’s stringent requirements for rigor, a LeanGeo-proof must explicitly account for all possible cases. While this often results in more intricate proofs, it also ensures a higher level of rigor compared with natural language and other formal systems such as AlphaGeometry and TongGeometry. 
\vspace{-0.4cm}

\vspace{-0.3cm}
To avoid overly cumbersome case analyses, we make extensive use of SMT solvers in our formal proofs to simplify the classification process and trivial results.

\subsection{LeanSMT 4.15}

To efficiently discharge goals deemed trivial in natural language proofs, LeanGeo invokes the CVC5~\citep{barbosa2022cvc5} SMT solver. In LeanEuclid~\citep{murphy2024autoformalizing}, the SystemE axioms are embedded as hardcoded SMT commands, and a custom translator converts Lean’s tactic states into SMT goals. By contrast, LeanGeo employs the \verb|esmt| tactic, which directly passes all local hypotheses from the current tactic state—together with SystemE’s inference axioms and the negated goal—to CVC5 for an unsatisfiability check. If CVC5 returns \texttt{unsat}, the entailment is confirmed.

For performance optimization, raw axiom expressions are not repeatedly translated into SMT commands. Instead, parsed axiom expressions are cached, and a global metavariable (mvar) dependency graph is maintained. This graph is dynamically updated whenever a definition or axiom annotated with \verb|@[euclid]| is encountered as shown in Listing \ref{lst:euclid-example}. The core logic for updating this dependency graph is presented in the Appendix \ref{euclid_appendix}.

\begin{lstlisting}[caption={Example of @[euclid] tactic usage}, label={lst:euclid-example}, frame=single]
@[euclid]
axiom zero_segment_if :
  $\forall$ (a b : Point),  |(a - b)| = 0 → a = b
\end{lstlisting}



\vspace{-0.5cm}
The \verb|@[euclid]| tactic makes our system more extensible. In LeanEuclid, the translator does not natively handle new definitions, meaning it would require manual modification to work with non-SystemE definitions such as sin and cos. Our system is designed to seamlessly incorporate such new definitions, making it more adaptable to a wider range of geometric problems. In addition, our theorem library inherits the expression styles of other tactics from LeanEuclid, such as \verb|euclid_intros|, \verb|euclid_apply|, and \verb|euclid_finish|. When these tactics are executed, the system automatically invokes LeanSMT to return the results. The specific usage and examples of these tactics can be found in Appendix  \ref{Library}.
\section{LeanGeo-Bench}
\subsection{Benchmark}
LeanGeo-Bench is a formal benchmark tailored for formalizing and proving contest-level plane geometry theorems in Lean 4 and LeanGeo. As shown in Table~\ref{sample-table}, the benchmark consists of 122 problems drawn from diverse sources, including existing theorem libraries, textbooks, synthetically generated problems, contest problems.
\vspace{-0.2cm}
\begin{table}[h]
\caption{Composition of LeanGeo-Bench}
\label{sample-table}
\begin{center}
\begin{tabular}{lcll}
\multicolumn{1}{c}{\bf SECTION}  &\multicolumn{1}{c}{\bf N}
&\multicolumn{1}{c}{\bf SOURCE} &\multicolumn{1}{c}{\bf METHOD}\\ \hline \\
UniGeo(UG)        &10 & LeanEuclid & Manually Written\\ 
Library(LB) & 10 & LeanGeo Library & Manually Written \\
Synthetic Problem(SP) & 20 & LeanGeo Library & Generated by gemini \\
High Shool Competition(HSC) & 20 & NuminaMath & Autoformalized + double check\\
Olympic Problem(OP) & 19 & Evan Chen's textbook & Autoformalized + double check \\
IMO & 43 & AoPS & Autoformalized + double check \\

\end{tabular}
\end{center}
\end{table}

\vspace{-0.2cm}
The benchmark's difficulty ranges from foundational to competition-level. It includes 20 introductory problems: 10 from UniGeo\citep{chen2022unigeo} and 10 from LeanGeo theorem library. Another 20 problems (`Gemini\_synthetic') are synthetically generated by an gemini-2.5 via our Problem Generation Pipeline. The majority of the benchmark consists of 83 more advanced problems sourced from high-school curricula, NuminaMath\citep{li2024numinamath}, Evan Chen's Geometry textbook \cite{chen2021euclidean}, and all the International Mathematical Olympiad (IMO) geometry problems since 2000 from AoPS\citep{AoPS:site}. These problems were developed using a human-in-the-loop methodology: For each problem, it is first autoformalized by a large language model through prompt engineering, and then rigorously reviewed and corrected by two human experts.

\begin{figure}[h]
    \centering
    \includegraphics[width=0.5\linewidth]{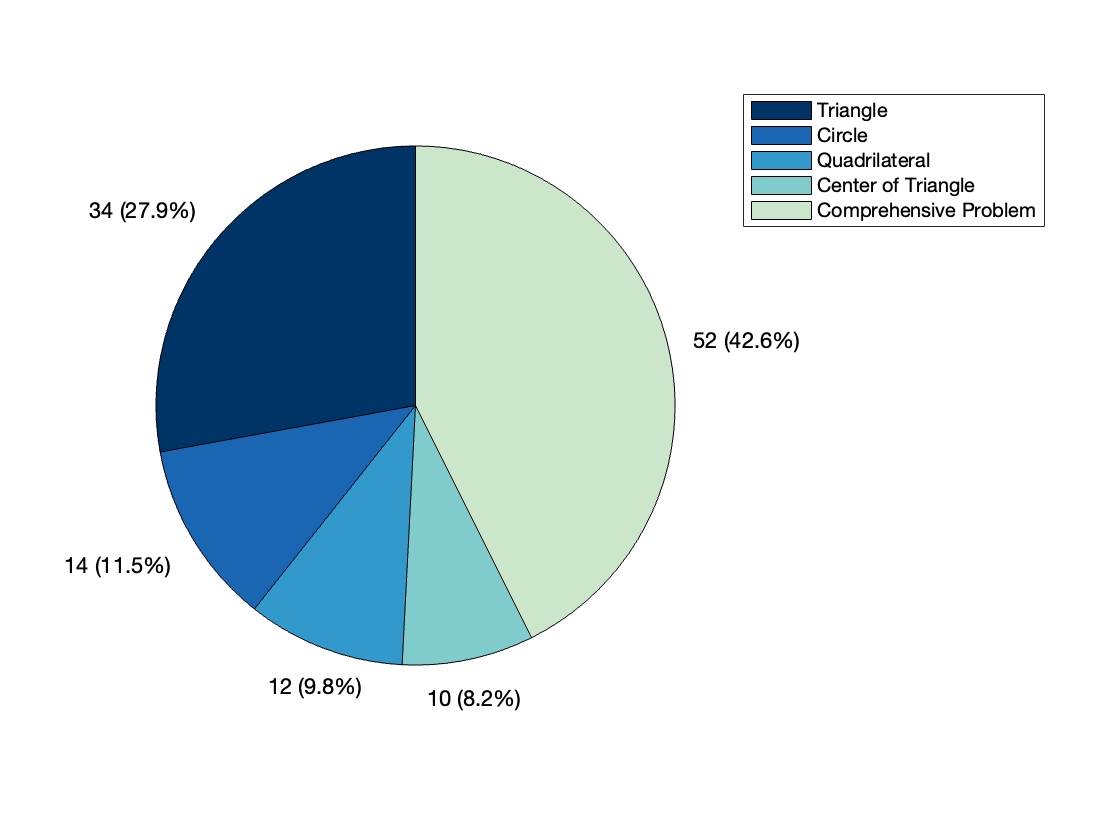}
    \vspace{-0.9cm}
    \caption{Category Distribution of LeanGeo-Bench}
    \label{fig:distribution}
\end{figure}

The benchmark covers a broad range of topics commonly encountered in competitive geometry, including triangles, circles, quadrilaterals, and notably triangle centers (e.g., incenter, circumcenter), as shown in Figure \ref{fig:distribution}. It also contains comprehensive problems involving multiple geometric configurations. Moreover, the problem types are diverse: in addition to traditional plane geometry proofs, many problems require calculating or deriving angles and side lengths. The benchmark further includes three geometry inequality problems and two problems involving moving points.

As part of this work, we present 43 formally verified solutions to problems in the benchmark, including two from the International Mathematical Olympiad (IMO), all of which are machine-checked in Lean. The formal proofs ensure the correctness of these problems. For problems without formal proofs, we validate correctness using a negation-based method combined with independent reviews by two geometry experts.

\subsection{Evaluation Method}
To guide the LLM in generating formal proofs, we design a comprehensive prompt that carefully structures the task environment. The prompt comprises a custom declarative Domain-Specific Language of LeanGeo, ``Error-and-correction" examples, construction rules for geometric definitions, the full set of theorems from the LeanGeo theorem library, together with few-shot learning examples. The complete prompt is provided in the Appendix \ref{Appendix:Prompt}.

To evaluate the result generated by LLM, we apply the \texttt{online\_one\_stage} Fine-Eval method introduced in CombiBench~\citep{liu2025combibench} -  This evaluation followed a two-step procedure. First, we checked that the LLM's result was consistent with the initial formal problem statement. Then, we fed the result into a Lean server containing a pre-built theorem library to formally verify the proof.

\subsection{Baseline Result}
To comprehensively evaluate the model's performance on the benchmark, we conducted extensive testing across Gemini 2.5 Pro~\citep{deepmind2025gemini2.5pro}, o4-mini~\citep{openai2025o3o4mini}, Grok 4~\citep{xai2025grok4} , Kimi K2~\citep{moonshotai2025kimiK2} ,  Claude 4~\citep{anthropic2025claude4} and Qwen3-235B-A22B~\citep{yang2025qwen3} and collected their overall success rates at different sample budgets and their performance in different section. The results are shown in Table \ref{eval-result}.
\vspace{-0.5cm}
\begin{table}[h]
\caption{Evaluation on LeanGeo-Bench}
\label{eval-result}
\begin{center}
\begin{tabular}{lccccccccc}

\multicolumn{1}{c}{\textbf{MODEL}} &\multicolumn{3}{c}{\textbf{OVERALL SUCCESS RATE (\%)}} & \multicolumn{6}{c}{\textbf{SUCCESS NUMBER(pass@4)}} \\ \cline{2-10}
 & pass@1 & pass@2 & pass@4 & UG & LB & SP & HSC & OP &IMO\\ \hline 
 Gemini 2.5 Pro & 17.21 & \textbf{22.95} & \textbf{27.05} & \textbf{10}& 4 & \textbf{13} & \textbf{6} & 0 & 0  \\
 o4-mini  & \textbf{19.67} & 21.31 & 22.13 & 7 & \textbf{9} & 8 & 3 & 0 & 0  \\
Grok 4  &  16.39 & 21.31 & 24.59  & \textbf{10} & 6 & 11 & 3 & 0 & 0\\
Kimi K2 & 9.02 & 9.02 & 9.84 & 1 & \textbf{9} & 2 & 0 & 0 & 0 \\
Claude 4 & 4.92 & 9.02 & 10.66 & 1 &5 & 7 & 0 & 0 & 0 \\
Qwen3-235B-A22B & 3.28  & 4.10 & 5.74 & 0 &6 &1 &0 & 0 & 0 \\
\hline
 & & & Total& 10 &10 & 20 &20 &19 & 43 \\
\end{tabular}
\end{center}
\end{table}

\vspace{-0.7cm}
The LeanGeo-Bench results reveal substantial differences in geometric theorem-proving performance across state-of-the-art LLMs. o4-mini~\citep{openai2025o3o4mini} attains the highest pass@1 score (19.67\%), while Gemini 2.5 Pro~\citep{deepmind2025gemini2.5pro} leads at pass@4 (27.05\%). 

A breakdown by category at pass@4 reveals complementary strength of LLMs in different area: Gemini-2.5-Pro excels in novel-problem settings such as Synthetic Proof (SP) and High School Competition (HSC), indicating stronger adaptability to unseen reasoning patterns, while GPT-o4-mini demonstrates greater proficiency in Library(LB), suggesting a more understanding and application of the  theorem library in prompt. 

While most models achieve partial success on the benchmark, their performance plateaus below 30\%, and notably none of the evaluated models could solve any of the 62 Olympic-level problems, indicating fundamental limitations in handling complex geometric proofs that require sophisticated logical reasoning, advanced diagram interpretation, and formal verification capabilities.

\section{Reinforcement Learning Experiments}
\subsection{Generating data by LLM}
A significant challenge in applying Reinforcement Learning training on LeanGeo is the absence of pre-existing cold start data, as LeanGeo establishes a novel framework for formal geometry. To address this, we developed a synthetic data generation pipeline. This process begins by creating a specialized prompt for Gemini 2.5 Pro~\citep{deepmind2025gemini2.5pro}, featuring carefully crafted guidelines and few-shot examples of theorem generation. Instead of tasking the LLM with solving a predefined problem, we prompt it with five randomly sampled theorems from our existing LeanGeo library. The LLM is then instructed to synthesize a new theorem and a corresponding proof, using the sampled theorems as inspiration. We repeated this process 5,000 times, each time conditioning the model on a different random subset of our library, to ensure a broad and diverse distribution of new problems.

\begin{figure}[h]
    \centering
    \includegraphics[width=0.8\linewidth]{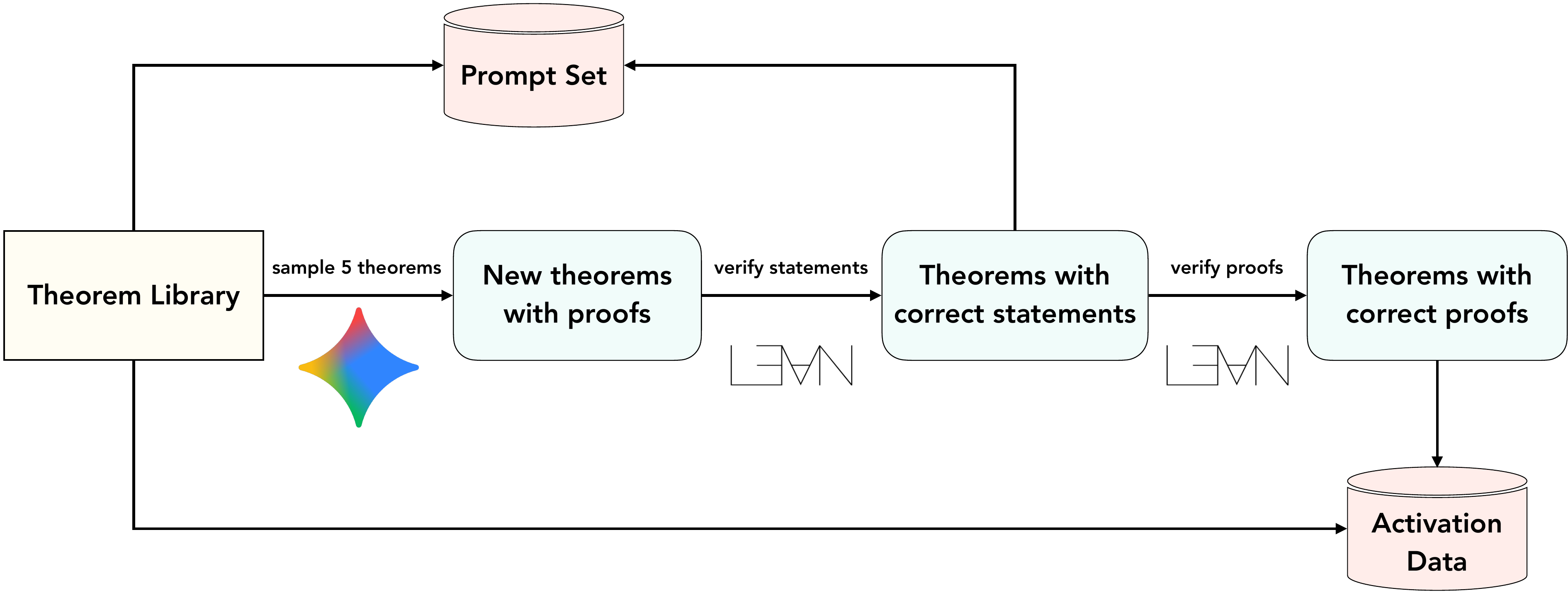}
    \caption{Data Generation Pipeline}
    \label{fig:RL}
\end{figure}

The generated theorem-proof pairs are then automatically verified using the Lean prover. This verification reveals that 89\% of the generated formal statements are syntactically valid, and 14\% of the full submissions (statement and proof) pass the verification. Based on this outcome, we categorize the generated data: the activation dataset consists of problems with a valid statement and correct proof. This dataset is used for supervised fine-tuning as the initialization phase for reinforcement learning, while problems with valid statement but invalid proof are used for the prompt set in reinforced learning. The whole process is illustrated in Figure \ref{fig:RL}.

\subsection{Instilling Knowledge in RL}
Another challenge arises from the size of our theorem library. To prove a new theorem, the model must select and apply relevant theorems from this library. Incorporating the entire library into the prompt may present practical limitations, as it risks surpassing the model’s context window, which could adversely impact training efficiency and model performance. To overcome this, we propose an ``instilling method" that structures the prompt to manage the context effectively. Specifically, we use the following data format:
\begin{lstlisting}[frame = single]
You are an expert in Lean 4 and geometric problem-solving.
You may apply the following theorems to solve the problem:
<theorem_1>
<theorem_2>
...
<theorem_10>
Now, let's solve the following problem step-by-step.
<formal_statement>
\end{lstlisting}

During reinforcement learning, we retain the same prompt structure; however, the 10 provided theorems are selected entirely at random from the library, regardless of their relevance to the target formal statement. This approach encourages the model to discern and apply theorems that are truly pertinent within a noisy context, fostering a critical skill necessary for effective problem-solving.

\subsection{RL Training}
We employ the RL framework of the Kimina-Prover~\citep{wang2025kimina} to train our model. Our RL training procedure consists of two stages. Initially, the agent is trained on the activation dataset, during which the model’s proof success rate improves from a post-SFT baseline of 37\% to 60\%. Subsequently, training proceeds on the prompt set, where the success rate increases from 12.5\% to 40\%. This training regimen also yields enhanced performance on our evaluation benchmark, with the pass@1 rate rising from 2.52 \% to 10.92\%.

\section{Discussion and Future Work}
While LeanGeo successfully demonstrates the viability of a declarative, human-readable approach for competition-level geometry, several key challenges and opportunities for future work remain. These are centered on strengthening the system's foundational soundness, enhancing its automation capabilities, and Instilling domain-specific knowledge to LLMs.

\subsection{Soundness}
Currently, the \verb|esmt| tactic relies on proof certificates generated by the external CVC5 SMT solver. For full trust and verification within Lean's kernel, these certificates must be converted into a native Lean proof. Although LeanSMT supports this functionality, we have encountered technical challenges with proof reconstruction after integrating a custom caching mechanism for axioms.
Therefore, a primary area for future development is ensuring the complete, end-to-end soundness of all proofs. 

\subsection{Automation Capabilities}
While the integration with SMT solvers is powerful, a limitation of general-purpose SMT solvers is their lack of geometry-specific heuristics. Therefore, the solving speed of SMT significantly decreases as the number of points in the problem increases. One way to scale LeanGeo for more complex problems is by embedding domain-specific proof automation, like the Area Method\citep{janicic:hal-00426563} or algebraic geometry techniques, into the tactic framework.


\subsection{Instilling Domain-Specific Knowledge to LLMs}
In the current benchmark, to ensure the model correctly cites theorems, we input the entire theorem library's statements as prompts to the model. However, long prompts may negatively impact the model's performance.

To address this issue, our RL framework takes first steps in reducing prompt length and instilling knowledge into LLMs. However, our method is still rather rudimentary and needs more sophisticated development.

\section{Conclusion}
In this paper, we present LeanGeo, the first Lean-based framework capable of formalizing and solving competition-level geometry problems, together with LeanGeo-Bench, a 122-problem benchmark spanning from foundational theorems to IMO challenges. LeanGeo’s declarative, human-readable proofs, deep Mathlib integration, and extensible library enable rigorous cross-domain reasoning beyond the reach of existing geometry systems.

Our baseline evaluations reveal that while current LLMs can solve some problems, they fall far short on the hardest tasks, underscoring the need for stronger geometric reasoning and proof search capabilities. By combining a rich formal library, a challenging benchmark, and initial reinforcement learning experiments, LeanGeo establishes a scalable testbed for advancing automated geometry theorem proving and neuro-symbolic reasoning.

\bibliography{iclr2026_conference}
\bibliographystyle{iclr2026_conference}

\newpage

\appendix
\section{Command Caching}
\label{euclid_appendix}
\begin{figure}[htbp]
\centering
\begin{lstlisting}[frame=single]
/--
Adds a command for a new constant to the SMT command cache and updates the dependency graph.

* `oldAxiomExprs`: the expressions corresponding to the types of all currently cached axioms.
* `cName`: the name of the axiom to be added to the cache.
* `initialState`: the current state of the global dependency graph.

Returns a tuple of the form `(new global dependency graph, new list of cached axioms, list of SMT commands for all of the axioms)`.
-/
def addCommandForConstant
  (oldAxiomExprs : List Expr)
  (cName : Name)
  (initialState : QueryBuilderM.State)
  : MetaM (QueryBuilderM.State × List Expr × List Command) := do
  let constInfo ← getConstInfo cName
  let constExpr := mkConst cName (constInfo.levelParams.map Level.param)
  let ((_, st), r) ←
    QueryBuilderM.buildDependencyGraph (mkConst `True)
    |>.run { toDefine := oldAxiomExprs ++ [constExpr] : QueryBuilderM.Config }
    |>.run initialState
    |>.run { uniqueFVarNames := {} : TranslationM.State }
  let (_, cmds) ← StateT.run (st.graph.orderedDfs (oldAxiomExprs ++ [constExpr]) (emitVertex st.commands)) []
  return ⟨st, oldAxiomExprs ++ [constExpr], cmds⟩
\end{lstlisting}
\caption{Command caching code for SystemE axioms.}
\end{figure}

\section{Changes to SystemE formalism}
\label{Appendix:Change}
There are some descrepencies between how SystemE axioms are described in the LeanEuclid lean theory vs how they are passed into the SMT solver. In particular \verb|degree| and \verb|length| and \verb|area| are defined directly as functions from Points to a real number. That is the types \verb|Angle| and \verb|Segment| do not exist in the SMT query. If a rule involves substituting a function into application into a forall statement it will double the search depth required to obtain that proof. For example if angle degree is defined as \verb|Angle.degree (Angle.ofPoints a b c)| the smt's search procedure would have to first apply Angle.ofPoints to points $a, b, c$ and then apply Angle.degree to that resultant angle. By contrast, if degree is defined as the measure of three points only a single application is required to obtain the term \verb|degree a b c|. By changing the definition of degree to be a function on three points it halves the search depth required to acheive the same term. Since we generally never reason about segments or angles outside of their measures this simplification is acceptable and segment congruence is defined uniquely by length. For Triangles it is not possible to get rid of the type entirely since Triangle congruence. We can however define a function \verb|area'| which behaves as an area function on points. When then define \verb|Triangle.area| (Triangle.ofPoints a b c) = area' a b c. And tag it as a simp lemma. Thus, since simplification is applied before passing into the smt solver, the Triangle type will dissappear by the time the smt solver is invoked. A similar trick can be done Triangle.congruence.

\begin{lstlisting}[frame=single,caption=Angle Definition]
opaque Angle : Point → Point → Point → ℝ
-- ...
notation:71 "$\angle$" a ":" b ":" c:72 => Angle a b c
\end{lstlisting}

\begin{lstlisting}[frame=single,caption=Triangle Definition]
opaque area' : Point → Point → Point → ℝ

inductive Triangle
| ofPoints (a b c : Point)

@[simp]
abbrev Triangle.area : Triangle → ℝ :=
  fun x =>
    match x with
    | ofPoints a b c => area' a b c


notation:max "$\triangle$" a ":" b ":" c:66 => Triangle.ofPoints a b c

instance : Coe Triangle ℝ :=
  ⟨Triangle.area⟩
\end{lstlisting}


Besides, to broaden SystemE’s applicability to the wider field of geometry, we add nine axioms to LeanGeo covering circles, triangles, similar triangles, and triangle areas, which cannot be derived within the original SystemE.

\begin{lstlisting}[frame=single,caption = Additional Axioms in LeanGeo]
axiom triangle_area_foot :∀ (a b c d: Point) (BC: Line),b.onLine BC ∧ c.onLine BC ∧ (Triangle a b c) ∧ Foot a d BC → ($\triangle$ a:b:c).area = |(a-d)| * |(b-c)|/2

axiom threePoints_existCircle : ∀ (A B C : Point),
  Triangle A B C →
  ∃ (Ω : Circle),
    (A.onCircle Ω ∧ B.onCircle Ω ∧ C.onCircle Ω)

axiom exists_centre : ∀ (O: Circle), ∃ (C : Point), C.isCentre O

axiom rightAngle_eq_pi_div_two : $\RightAngle$ = Real.pi / 2

axiom rightTriangle_sin : ∀ (A B C : Point), RightTriangle A B C → Real.sin ($\angle$A:B:C) = |(A-C)| / |(B-C)|

axiom rightTriangle_cos : ∀ (A B C : Point), RightTriangle A B C → Real.cos ($\angle$A:B:C) = |(A-B)| / |(B-C)|

axiom similar_AA : ∀ (A B C D E F : Point), Triangle A B C ∧ Triangle D E F ∧  $\angle$ A:B:C = $\angle$ D:E:F ∧ $\angle$ B:A:C = $\angle$ E:D:F → SimilarTriangles A B C D E F

axiom similar_SAS : ∀ (A B C D E F : Point), Triangle A B C ∧ Triangle D E F ∧  $\angle$ A:B:C = $\angle$ D:E:F ∧ |(A-B)| * |(E-F)| = |(B-C)| * |(D-E)| → SimilarTriangles A B C D E F

axiom similar_SSS : ∀ (A B C D E F : Point), Triangle A B C ∧ Triangle D E F ∧ |(A-B)| * |(E-F)| = |(B-C)| * |(D-E)| ∧ |(B-C)| * |(F-D)| = |(C-A)| * |(E-F)| → SimilarTriangles A B C D E F
    
\end{lstlisting}

\section{Examples of Formalization}

\subsection{Examples in Theorem Library}
\label{Library}
Here is a proof example from the LeanGeo theorem library. 
\begin{lstlisting}[caption={Example of Theorem Library},label={example_library},frame=single]
theorem angle_lt_outsideCircle: ∀ (A B C D : Point) (AB : Line) (Ω : Circle), A.onCircle Ω ∧ B.onCircle Ω ∧ distinctPointsOnLine A B AB ∧ C.onCircle Ω ∧ C ≠ A ∧ C ≠ B ∧ D.sameSide C AB ∧ $\angle A:D:B$ < $\angle$ A:C:B → D.outsideCircle Ω := by
  euclid_intros
  have h1 : ¬ (D.onCircle Ω) := by
    by_contra
    euclid_apply cyclic_eqAngle A B C D AB Ω
    euclid_finish
  have h2: ¬ (D.insideCircle Ω):= by
    by_contra
    euclid_apply line_from_points A D as AD
    euclid_apply intersection_circle_line_extending_points Ω AD D A as E
    have h3: $\angle$ B:C:A = $\angle$ B:E:A := by
      euclid_apply cyclic_eqAngle A B C E AB Ω
      euclid_finish
    euclid_apply triangle_exteriorAngle E D B A
    have h4: $\angle$ A:E:B = $\angle$ D:E:B := by
      euclid_apply angle_between_transfer A D E B
      euclid_finish
    euclid_finish
  euclid_finish
\end{lstlisting}
LeanGeo proofs are structured to mirror the step-by-step, declarative style of traditional, natural-language geometry proofs. This design choice results in simple, readable proof scripts that are particularly amenable to machine learning techniques. The proof development relies on a small set of core tactics:
\begin{itemize}
    \item \verb|euclid_intros| \\
    This is an initialization tactic that begins the proof. It processes the theorem's statement, automatically introducing all universally quantified variables (e.g., `A', `B', `C', `D', `$\Omega$') and hypotheses (e.g., `A.onCircle $\Omega$', `D.sameSide C AB') into the local proof context.
    \item \verb|euclid_apply <rule> <args>|  \\
    Given a rule \verb|<rule>| with type of the form \texttt{$\forall$(<args> : Types) ... P -> Q}, this tactic attempts to prove premise P from the local proof and attempts to prove premise P from the local proof context using an SMT solver. If successful, propsition Q is added to the proof context.
    
    In this example, \texttt{euclid\_apply cyclic\_eqAngle A B C D AB} refers to the former theorem in the library(in Circle.lean)
    
\begin{lstlisting}[frame=single]
    theorem cyclic_eqAngle: ∀ (A B C D: Point) (AB:Line) (Ω : Circle), distinctPointsOnLine A B AB ∧ C≠ A ∧ D ≠ A ∧ C ≠ B ∧ D ≠ B ∧ A.onCircle Ω ∧ B.onCircle Ω ∧  C.onCircle Ω ∧ D.onCircle Ω ∧ C.sameSide D AB → $\angle$ B:C:A = $\angle$ B:D:A := by ...
\end{lstlisting}
    LeanGeo automatically checks whether all of the premises of \texttt{cyclic\_eqAngle}, i.e. \texttt{distinctPointsOnLine A B AB, C $\neq$ A, D $\neq$ A ...} are satisfied. If yes, then its result,\texttt{$\angle$ B:C:A = $\angle$ B:D:A} will be added in the proof context.
    \item   \texttt{euclid\_apply <rule> with <args> as <x, h>} \\
    A forward-reasoning tactic designed to apply theorems and construction rules. Given a rule, typically of the form \texttt{$\forall$..., P → $\exists$ x, Q(x)} \\
    This tactic instantiates it with the provided arguments \texttt{<args>}. It then employs an SMT solver to automatically prove the premise `P` using hypotheses from the local context. If successful, the tactic introduces the newly constructed object 'x' and its property 'Q(x)' (named 'h') into the context. This command streamlines geometric constructions and deductions by combining the application of a rule with the automated verification of its preconditions, making the proof script more declarative and readable.
    \item  \texttt{euclid\_finish}\\
    A terminal tactic that invokes an SMT solver to automatically prove the current goal using the set of available hypotheses in the local context. This tactic is effective for discharging goals that are either direct assumptions or straightforward logical consequences of the premises, requiring minimal search from the solver.
    \item \texttt{have hP : P := by}  \\
    A construct for structuring proofs by introducing an intermediate lemma `P` (named `hP`). This allows a complex proof to be decomposed into a sequence of smaller, more manageable sub-proofs. This methodology not only enhances the readability and maintainability of the proof script but also improves the SMT solver's performance by reducing its search space. The solver can tackle the smaller lemma in isolation and then utilize the proven result `hP` in the main proof.
    \item

\end{itemize}
\subsection{Formalization of IMO 2001 P1}
\label{Appendix:imo2001}
\lstset{
  basicstyle=\ttfamily\small,
  frame=single,                
  frameround=tttt,             
  rulecolor=\color{gray},      
}
Problem statement:
\begin{lstlisting}[frame=single]
Let $A B C$ be an acute-angled triangle with $O$ as its circumcenter. Let $P$ on line $B C$ be the foot of the altitude from $A$. Assume that $\angle B C A \geq \angle A B C+30^{\circ}$. Prove that $\angle C A B+\angle C O P<90^{\circ}$. 
\end{lstlisting}

Proof of LeanGeo:
\begin{lstlisting}[caption=Proof of LeanGeo for IMO 2001 P1]
import Mathlib
import SystemE
import LeanGeo
open LeanGeo Real
--Consider an acute-angled Triangle $ABC$. Let $P$ be the Foot of the altitude of Triangle $ABC$ issuing from the vertex $A$, and let $O$ be the circumcenter of Triangle $ABC$. Assume that $\angle C \geq \angle B+30^{\circ}$. Prove that $\angle A+\angle COP < 90^{\circ}$.
--To Trigonometry.lean
--To Triangle.lean
set_option maxHeartbeats 0

theorem sin_inequality(B C : ℝ)
  (hB : 0 < B ∧ B < π) (hC : 0 < C ∧ C < π)
  (hC1 : C ≥ B + π/6) : 4 * sin B * cos C ≤ 1 := by
  rcases hB with ⟨hB1, hB2⟩
  rcases hC with ⟨hC11, hC22⟩
  have h1 : cos C ≤ cos (B + π / 6) := by
    have h2 : C ≥ B + π / 6 := hC1
    have h3 : C < π := by linarith [hC22]
    have h4 : 0 < B + π / 6 := by
      linarith [hB1, Real.pi_pos]
    have h5 : B + π / 6 < π := by
      nlinarith [hB2, hC11, hC22, Real.pi_pos]
    have h6 : cos C ≤ cos (B + π / 6) := by
      apply Real.cos_le_cos_of_nonneg_of_le_pi
      all_goals
        nlinarith [Real.pi_pos, hB1, hB2, hC11, hC22, Real.pi_pos]
    linarith
  have h2 : sin B * cos (B + π / 6) ≤ 1 / 4 := by
    have h21 : cos (B + π / 6) = cos B * cos (π / 6) - sin B * sin (π / 6) := by
      rw [Real.cos_add]
    have h22 : cos (π / 6) = Real.sqrt 3 / 2 := by
      rw [cos_pi_div_six]
    have h23 : sin (π / 6) = 1 / 2 := by
      rw [sin_pi_div_six]
    have h24 : sin B * cos (B + π / 6) = (Real.sqrt 3 / 2) * sin B * cos B - (1 / 2) * sin B ^ 2 := by
      rw [h21, h22, h23]
      ring_nf
    have h25 : (Real.sqrt 3 / 2) * sin B * cos B - (1 / 2) * sin B ^ 2 ≤ 1 / 4 := by
      nlinarith [sq_nonneg (sin B - 1 / 2), sq_nonneg (cos B - Real.sqrt 3 / 2),
          sq_nonneg (sin B ^ 2 - 1 / 4), sq_nonneg (sin B - Real.sqrt 3 / 2),
          sq_nonneg (cos B ^ 2 - 1 / 4), sq_nonneg (cos B - 1 / 2),
          Real.sqrt_pos.mpr (by linarith : (0 : ℝ) < (3 : ℝ)),
          Real.sqrt_nonneg 3, Real.sq_sqrt (show (0 : ℝ) ≤ (3 : ℝ) by linarith),
          Real.sin_sq_add_cos_sq B, mul_nonneg (show 0 ≤ (0 : ℝ) by linarith) (show 0 ≤ (0 : ℝ) by linarith),
          Real.sin_pos_of_pos_of_lt_pi hB1 (by linarith : B < Real.pi)]
    linarith [h24, h25]
  have h3 : 0 < sin B := by
    apply sin_pos_of_pos_of_lt_pi
    all_goals linarith [hB1, hB2, Real.pi_pos]
  nlinarith [h1, h2, h3, Real.sin_sq_add_cos_sq B, Real.sin_sq_add_cos_sq C, Real.pi_pos]


theorem sin_range (A : ℝ) (hA : 0 < A ∧ A < π/2) : sin A < 1 ∧ sin A > 0 := by
  have h1 : 0 < A := hA.1
  have h2 : A < π / 2 := hA.2
  have h3 : sin A < 1 := by
    have h4 : sin (π / 2) = 1 := by
      rw [sin_pi_div_two]
    have h5 : sin A < sin (π / 2) := by
      apply sin_lt_sin_of_lt_of_le_pi_div_two
      all_goals linarith [Real.pi_pos, Real.pi_gt_three, h1, h2]
    linarith [h4, h5]
  have h6 : sin A > 0 := by
    have h7 : sin (0 : ℝ) = 0 := by
      simp [Real.sin_zero]
    have h8 : sin (0 : ℝ) < sin A := by
      apply sin_lt_sin_of_lt_of_le_pi_div_two
      all_goals linarith [Real.pi_pos, Real.pi_gt_three, h1, h2]
    linarith [h7, h8]
  constructor
  · linarith [h3]
  · linarith [h6]
--To Triangle, Generated b


theorem IMO_2001_P1 :
  ∀ (A B C P O : Point) (AB BC CA : Line),
    formAcuteTriangle A B C AB BC CA ∧
    Foot A P BC ∧
    Circumcentre O A B C ∧
    $\angle$ A:C:B ≥ $\angle$ C:B:A + $\RightAngle$/3 →
    $\angle$ B:A:C + $\angle$ C:O:P < $\RightAngle$ := by
  euclid_intros
  euclid_apply rightAngle_eq_pi_div_two
  euclid_apply acuteTriangle_circumcentre_insideTriangle A B C O AB BC CA
  euclid_apply circle_from_points O B as Ω
  euclid_apply circumcentre_inscribedAngle_comp B C A O BC Ω
  have h0: 4 * sin ($\angle$ B:A:C) * sin ($\angle$A:B:C) * cos ($\angle$A:C:B) < 1 := by
    have h1: 0 < $\angle$ A:B:C ∧ $\angle$ A:B:C < π := by
      euclid_finish
    have h2: 0 < $\angle$ A:C:B ∧ $\angle$ A:C:B < π := by
      euclid_finish
    have h3: (sin ($\angle$ B:A:C) < 1) ∧ (sin ($\angle$ B:A:C) > 0) := by
      euclid_apply sin_range ($\angle$B:A:C)
      euclid_finish
    have h4: $\angle$ A:C:B ≥ $\angle$ C:B:A + π/6 := by
      euclid_finish
    have h5: 4 * sin ($\angle$A:B:C) * cos ($\angle$A:C:B) ≤ 1 := by
      euclid_apply sin_inequality ($\angle$A:B:C) ($\angle$A:C:B)
      euclid_finish
    nlinarith

  have h1: between B P C := by
    euclid_apply acuteTriangle_foot_between A B C P BC
    euclid_finish
  have h2: |(P-C)| < |(P-O)| := by
    have h3: |(P-C)| * |(P-C)|  < |(P-O)| * |(P-O)| := by
      have h4: |(O-C)| * |(O-C)| - |(O-P)| * |(O-P)| = |(P-B)| * |(P-C)|:= by
        euclid_apply ApolloniusTheorem_to_isoTriangle O B C P BC
        euclid_finish
      have h5: |(P-C)| = |(A-C)| * cos ($\angle$ A:C:P) := by
        euclid_apply rightTriangle_cos P C A
        euclid_finish
      have h6:  |(A-C)| = 2 * |(O-C)| * sin ($\angle$A:B:C) := by
        euclid_apply LawOfSines_radius B A C O
        euclid_finish
      have h7:  |(B-C)| = 2 * |(O-C)| * sin ($\angle$B:A:C) := by
        euclid_apply LawOfSines_radius A B C O
        euclid_finish
      have h8: $\angle$A:C:P = $\angle$A:C:B := by
        euclid_apply coll_angles_eq B P C A
        euclid_finish
      have h9: |(P-C)| * |(B-C)| < |(O-C)| * |(O-C)| := by
        rw [h5, h6, h7,h8]
        have h10: (|(O-C)| * |(O-C)|) > 0 := by euclid_finish
        calc
          _ = (4 * sin ($\angle$ B:A:C) * sin ($\angle$A:B:C) * cos ($\angle$A:C:B)) * (|(O-C)| * |(O-C)|) := by linarith
          _ < 1 * (|(O-C)| * |(O-C)|) := by euclid_finish
          _ = _ := by euclid_finish
      euclid_finish
    euclid_assert |(P-C)| > 0
    euclid_assert |(P-O)| > 0
    nlinarith
  euclid_assert Triangle O C P
  euclid_apply triangle_gt_side_gt_angle P C O
  have h_final: $\angle$ P:C:O = $\angle$ B:C:O := by
    euclid_apply coll_angles_eq B P C O
    euclid_finish
  euclid_finish
\end{lstlisting}
A significant advantage of LeanGeo is its seamless integration with Mathlib's extensive mathematical library, enabling it to tackle a broader class of problems . This is particularly evident in its ability to formalize geometric inequalities, a domain where systems like AlphaGeometry face challenges due to their reliance on converting geometry into polynomial equations. The formalization of IMO 2001 P1, shown above, serves as a prime example. The proof strategy involves reducing the geometric inequality $\angle CAB + \angle COP < \frac{\pi}{2}$ to a trigonometric one: $4 \sin(\angle ABC)  \cos(\angle BCA) \leq  1$, derived from the condition $\angle BCA \geq \angle ABC + \frac{\pi}{6} $.

This trigonometric lemma, `sin\_inequality', is proven not by geometric tactics. Annotators could obtain the proof from a open-sourced formal prover, Kimina-Prover \cite{wang2025kimina}. The main geometric proof, orchestrated by LeanGeo's `euclid\_...` tactics, then imports and applies this analytical result to complete the formalization. This hybrid approach, combining high-level geometric reasoning with deep analytical capabilities from Mathlib, demonstrates LeanGeo's power in unifying different mathematical domains to expand the scope of automated geometric theorem proving.

\subsection{Comparison of proof with AlphaGeometry of IMO 2000 P1}
\label{Appendix:Comparison}
We use IMO 2000 Problem 1 to illustrate LeanGeo's formalization of a geometric proof, allowing for a direct comparison with AlphaGeometry's solution.
\begin{lstlisting}[caption=IMO 2000 Problem 1,frame=single]
Two circles $G_1$ and $G_2$ intersect at two points $M$ and $N$. Let $AB$ be the line tangent to these circles at $A$ and $B$, respectively, so that $M$ lies closer to $AB$ than $N$. Let $CD$ be the line parallel to $AB$ and passing through the point $M$, with $C$ on $G_1$ and $D$ on $G_2$. Lines $AC$ and $BD$ meet at $E$; lines $AN$ and $CD$ meet at $P$; lines $BN$ and $CD$ meet at $Q$. Show that $EP=EQ$.
\end{lstlisting}
AlphaGeometry presents the proof as a flat, linear sequence of 44 atomic deductions. While logically sound, this format obscures the underlying geometric narrative. It reads as a symbolic log where high-level concepts, without explicitly grouping these steps into a coherent subgoal.
\begin{lstlisting}[caption=Proof of AlphaGeometry for IMO 2000 Problem 1,frame=single]
* Formal statement:
a b = segment a b; c = on_tline c a a b; d = on_tline d b b a; e = on_circle e c a, on_circle e d b; f = on_circle f c a, on_circle f d b; g = on_pline g e a b, on_circle g c a; h = on_pline h e a b, on_circle h d b; i = on_line i a g, on_line i b h; j = on_line j a f, on_line j g h; k = on_line k b f, on_line k g h ? cong i j i k
==========================
 * From theorem premises:
A B C D E F G H I J K : Points
AC $\perp$ AB [00]
BA $\perp$ DB [01]
DE = DB [02]
CE = CA [03]
DF = DB [04]
CF = CA [05]
$\angle$FAE = $\angle$FAE [06]
GE // AB [07]
CG = CA [08]
$\angle$GAF = $\angle$GAF [09]
HE // AB [10]
DH = DB [11]
$\angle$FBH = $\angle$FBH [12]
I,G,A are collinear [13]
I,B,H are collinear [14]
J,F,A are collinear [15]
J,G,H are collinear [16]
BF:BK = BF:BK [17]
G,K,H are collinear [18]
B,F,K are collinear [19]

 * Auxiliary Constructions:
: Points


 * Proof steps:
001. EG // AB [07] & EH // AB [10] ⇒  EH // EG [20]
002. EH // EG [20] ⇒  E,G,H are collinear [21]
003. DH = DB [11] & DF = DB [04] ⇒  D is the circumcenter of \Delta BHF [22]
004. D is the circumcenter of \Delta BHF [22] & DB $\perp$ BA [01] ⇒  $\angle$ABH = $\angle$BFH [23]
005. D is the circumcenter of \Delta BHF [22] & DB $\perp$ BA [01] ⇒  $\angle$ABF = $\angle$BHF [24]
006. E,G,H are collinear [21] & G,K,H are collinear [18] & $\angle$BFH = $\angle$ABH [23] & AB // EG [07] ⇒  $\angle$BFH = $\angle$KHB [25]
007. E,G,H are collinear [21] & G,K,H are collinear [18] & B,F,K are collinear [19] & $\angle$BHF = $\angle$ABF [24] & AB // EG [07] ⇒  $\angle$BHF = $\angle$HKB [26]
008. $\angle$BFH = $\angle$KHB [25] & $\angle$BHF = $\angle$HKB [26] (Similar Triangles)⇒  BF:BH = BH:BK [27]
009. DF = DB [04] & DH = DB [11] & DE = DB [02] ⇒  E,B,F,H are concyclic [28]
010. DF = DB [04] & DE = DB [02] ⇒  D is the circumcenter of \Delta BFE [29]
011. D is the circumcenter of \Delta BFE [29] & DB $\perp$ BA [01] ⇒  $\angle$EBA = $\angle$EFB [30]
012. E,G,H are collinear [21] & $\angle$EFB = $\angle$EBA [30] & AB // EG [07] ⇒  $\angle$EFB = $\angle$BEH [31]
013. E,B,F,H are concyclic [28] & $\angle$EFB = $\angle$BEH [31] ⇒  EB = BH [32]
014. CE = CA [03] & CG = CA [08] ⇒  C is the circumcenter of \Delta AEG [33]
015. C is the circumcenter of \Delta AEG [33] & AC $\perp$ AB [00] ⇒  $\angle$BAE = $\angle$AGE [34]
016. I,G,A are collinear [13] & $\angle$BAE = $\angle$AGE [34] & EG // AB [07] ⇒  $\angle$IAB = $\angle$BAE [35]
017. DH = DB [11] & DE = DB [02] ⇒  D is the circumcenter of \Delta BHE [36]
018. D is the circumcenter of \Delta BHE [36] & DB $\perp$ BA [01] ⇒  $\angle$ABH = $\angle$BEH [37]
019. I,B,H are collinear [14] & $\angle$ABH = $\angle$BEH [37] & EH // AB [10] ⇒  $\angle$ABE = $\angle$IBA [38]
020. $\angle$IAB = $\angle$BAE [35] & $\angle$ABE = $\angle$IBA [38] (Similar Triangles)⇒  BI = BE [39]
021. $\angle$IAB = $\angle$BAE [35] & $\angle$ABE = $\angle$IBA [38] (Similar Triangles)⇒  AI = AE [40]
022. BF:BH = BH:BK [27] & EB = BH [32] & BI = BE [39] ⇒  IB:BF = BK:IB [41]
023. B,F,K are collinear [19] & I,B,H are collinear [14] & $\angle$FBH = $\angle$FBH [12] ⇒  $\angle$KBI = $\angle$FBI [42]
024. IB:BF = BK:IB [41] & $\angle$KBI = $\angle$FBI [42] (Similar Triangles)⇒  BK:IK = IB:IF [43]
025. E,B,F,H are concyclic [28] ⇒  $\angle$FEH = $\angle$FBH [44]
026. CF = CA [05] & CG = CA [08] & CE = CA [03] ⇒  E,G,F,A are concyclic [45]
027. E,G,F,A are concyclic [45] ⇒  $\angle$GEF = $\angle$GAF [46]
028. I,G,A are collinear [13] & I,B,H are collinear [14] & $\angle$FEH = $\angle$FBH [44] & EH // AB [10] & $\angle$GEF = $\angle$GAF [46] & EG // AB [07] ⇒  $\angle$IAF = $\angle$IBF [47]
029. $\angle$IAF = $\angle$IBF [47] ⇒  I,B,F,A are concyclic [48]
030. I,B,F,A are concyclic [48] ⇒  $\angle$IBA = $\angle$IFA [49]
031. I,B,F,A are concyclic [48] ⇒  $\angle$IFB = $\angle$IAB [50]
032. E,G,H are collinear [21] & G,K,H are collinear [18] & J,F,A are collinear [15] & $\angle$IBA = $\angle$IFA [49] & I,B,H are collinear [14] & $\angle$ABH = $\angle$BEH [37] & EH // AB [10] & AB // EG [07] ⇒  $\angle$BEK = $\angle$JFI [51]
033. CE = CA [03] & CF = CA [05] ⇒  C is the circumcenter of \Delta AEF [52]
034. C is the circumcenter of \Delta AEF [52] & AC $\perp$ AB [00] ⇒  $\angle$BAE = $\angle$AFE [53]
035. J,G,H are collinear [16] & E,G,H are collinear [21] & $\angle$BAE = $\angle$AFE [53] & AB // EG [07] ⇒  $\angle$JEA = $\angle$AFE [54]
036. J,F,A are collinear [15] & $\angle$FAE = $\angle$FAE [06] ⇒  $\angle$JAE = $\angle$FAE [55]
037. $\angle$JEA = $\angle$AFE [54] & $\angle$JAE = $\angle$FAE [55] (Similar Triangles)⇒  JA:EA = EA:FA [56]
038. EA:FA = JA:EA [56] & IA = EA [40] ⇒  IA:FA = JA:IA [57]
039. I,G,A are collinear [13] & J,F,A are collinear [15] & $\angle$GAF = $\angle$GAF [09] ⇒  $\angle$IAF = $\angle$IAJ [58]
040. IA:FA = JA:IA [57] & $\angle$IAF = $\angle$IAJ [58] (Similar Triangles)⇒  $\angle$AIF = $\angle$IJA [59]
041. B,F,K are collinear [19] & E,G,H are collinear [21] & G,K,H are collinear [18] & J,F,A are collinear [15] & $\angle$AIF = $\angle$IJA [59] & I,G,A are collinear [13] & $\angle$IFB = $\angle$IAB [50] & AB // EG [07] ⇒  $\angle$BKE = $\angle$FJI [60]
042. $\angle$BEK = $\angle$JFI [51] & $\angle$BKE = $\angle$FJI [60] (Similar Triangles)⇒  BE:IF = BK:IJ [61]
043. BK:IK = IB:IF [43] & BE:IF = BK:IJ [61] & BI = BE [39] ⇒  BK:JI = BK:IK [62]
044. BF:BK = BF:BK [17] & BK:JI = BK:IK [62] ⇒  JI = IK
==========================
\end{lstlisting}
In contrast, the LeanGeo proof is structured more hierarchically, perfectly reflecting the problem’s intrinsic geometric structure. The proof is organized into clear, self-contained logical blocks, such as proving `midP\_ATB' (T is the midpoint of AB) or `perp\_EM\_CD'. Each block is achieved by invoking powerful theorems in LeanGeo library like `TangentSecantTheorem' and `AlternateSegmentTheorem' — mirroring the exact language a mathematicia would use. Consequently, the LeanGeo proof is not only verifiable but also intelligible, bridging the gap between a machine-generated proof trace and a human-authored mathematical argument. It demonstrates a system that reasons in a manner remarkably close to natural geometric intuition.
\begin{lstlisting}[caption=Proof of LeanGeo for IMO\_2000\_P1,frame=single]
import Mathlib
import SystemE
import LeanGeo
namespace LeanGeo
set_option maxHeartbeats 0
--To circle
--Two circles $G_1$ and $G_2$ intersect at two points $M$ and $N$. Let $AB$ be the line tangent to these circles at $A$ and $B$, respectively, so that $M$ lies closer to $AB$ than $N$. Let $CD$ be the line parallel to $AB$ and passing through the point $M$, with $C$ on $G_1$ and $D$ on $G_2$. Lines $AC$ and $BD$ meet at $E$; lines $AN$ and $CD$ meet at $P$; lines $BN$ and $CD$ meet at $Q$. Show that $EP = EQ$.
theorem IMO_2000_P1 :
  ∀ (M N A B C D E P Q O1 O2 : Point) (G1 G2 : Circle) (AB CD AC BD AN BN : Line),
    CirclesIntersectAtTwoPoints G1 G2 M N ∧
    distinctPointsOnLine A B AB ∧
    TangentLineCircleAtPoint A O1 AB G1 ∧
    TangentLineCircleAtPoint B O2 AB G2 ∧
    ¬ AB.intersectsLine CD ∧
    distinctPointsOnLine M C CD ∧
    C.onCircle G1 ∧ C ≠ M ∧ C ≠ N ∧
    D.onCircle G2 ∧ between C M D ∧
    distinctPointsOnLine A C AC ∧
    distinctPointsOnLine B D BD ∧
    between E A C ∧ between E B D ∧
    distinctPointsOnLine A N AN ∧
    TwoLinesIntersectAtPoint AN CD P ∧
    distinctPointsOnLine B N BN ∧
    TwoLinesIntersectAtPoint BN CD Q →
    |(E-P)| = |(E-Q)| := by
    euclid_intros
    euclid_apply line_from_points M N as MN
    euclid_apply intersection_lines MN AB as T
    have midP_ATB: MidPoint A T B := by
      have h1: |(T-A)| * |(T-A)| = |(T-M)| * |(T-N)| := by
        euclid_apply TangentSecantTheorem T A M N O1 G1 AB
        euclid_finish
      have h2: |(T-B)| * |(T-B)| = |(T-M)| * |(T-N)| := by
        euclid_apply TangentSecantTheorem T B M N O2 G2 AB
        euclid_finish
      have h3: |(T-A)| * |(T-A)| = |(T-B)| * |(T-B)| := by
        rw[h1,h2]
      euclid_assert |(T-A)| > 0
      euclid_assert |(T-B)| > 0
      have h4: |(T-A)| = |(T-B)| := by
        nlinarith
      euclid_finish
    have midP_PMQ : MidPoint P M Q := by
      have h1 : |(M-Q)| = |(M-P)| := by
        have h4: |(T-A)| = |(T-B)| := by euclid_finish
        have h5: |(M-Q)| * |(T-A)| = |(M-P)| * |(T-B)| := by
          euclid_apply triangle_parallel_bases_eq_ratio N T A M P B Q AB CD
          euclid_finish
        rw [h4] at h5
        have h6: |(T-B)| > 0 := by euclid_finish
        euclid_finish
      have h2: between P M Q := by
        euclid_finish
      euclid_finish
    euclid_apply line_from_points E M as EM
    have h_congr: CongruentTriangles A B E A B M := by
      have h1: $\angle$E:A:B = $\angle$M:A:B := by
        have h2: $\angle$E:A:B = $\angle$E:C:D := by
          euclid_apply parallel_imp_eq_alternateExteriorAngles B A D C E AB CD AC
          euclid_finish
        have h3: $\angle$M:A:B = $\angle$M:C:A := by
          euclid_apply line_from_points A M as AM
          have h4: M.sameSide B AC := by
            euclid_finish
          euclid_apply AlternateSegmentTheorem A M C B O1 G1 AM CD AC AB
          euclid_finish
        euclid_finish
      have h5: $\angle$E:B:A = $\angle$M:B:A := by
        have h6: $\angle$E:B:A = $\angle$E:D:C := by
          euclid_apply parallel_imp_eq_alternateExteriorAngles A B C D E AB CD BD
          euclid_finish
        have h7: $\angle$M:B:A = $\angle$M:D:B := by
          euclid_apply line_from_points B M as BM
          have h8: M.sameSide A BD := by
            euclid_finish
          euclid_apply AlternateSegmentTheorem B M D A O2 G2 BM CD BD AB
          euclid_finish
        euclid_finish
      euclid_apply congruentTriangles_ASA A B E A B M
      euclid_finish
    have perp_EM_CD: PerpLine EM CD := by
      have h1: PerpBisector E M AB := by
        euclid_apply perpBisector_if_eq_dist E M A B AB
        euclid_finish
      euclid_apply perpBisector_imp_perpLine E M EM AB
      euclid_apply perp_parallel_imp_perp AB EM CD
      euclid_finish
    have perpB: PerpBisector P Q EM := by
      euclid_apply (perpBisector_iff P Q EM).mpr
      euclid_finish
    euclid_finish

\end{lstlisting}
\section{Prompt for Evaluation}
\label{Appendix:Prompt}
\begin{lstlisting}[frame=single,caption=Prompt for LLMs in Evaluation]
    You are an expert of Lean 4. Now You are using a new Lean 4 system called LeanEuclid. The following is how you prove your theorem.
--- Proof DSL ---
Your proof must be a tactic proof in the LeanEuclid proof DSL. This DSL is built from
    the following tactics (arguments shown in angle-brackets <> ):
* TACTIC: euclid_intros *
 Introduces universally quantified variables and premises of the current goal into the proof context. No names required.
* TACTIC: euclid_apply <rule> <args> *
where <rule> is either a construction rule, inference rule, or other theorem.
Given a rule <rule> with type of the form ∀ (<args> : Types) ... P -> Q, this tactic
     instantiates <rule> with <args>, and attempts to prove premise P from the local proof
      context using an SMT solver. If successful, propsition Q is added to the proof
     context.
usage examples :
  euclid_apply PythagoreanTheorem_point a b c : SMT solver will try to search whether the premise of theorem "PythagoreanTheorem_point" i.e.(Triangle a b c) ∧ ($\angle$ b:a:c : ℝ) are satisfied, if not, the proof will fail. If all premises are found, then the conclusion of this theorem will be added to the solving context, i.e. |(b-c)| * |(b-c)| = |(b-a)| * |(b-a)| + |(a-c)| * |(a-c)|.

* TACTIC: euclid_apply <rule> <args> as X *
Given a rule <rule> with type of the form ∀ (<args> : Types) ... P -> ∃ x . Q(x), this
    tactic instantiates <rule> with <args>, and attempts to prove premise P from the local
     proof context using an SMT solver. If successful, object x and premise Q(x) are added
     to the proof context.
usage examples:
  euclid_apply line_from_points p1 p2 as M this tactic will first check whether p1 and p2 are different. If they are, then a new line M is added to the proof context and new condition, p1.onLine M and p2.onLine M will be added to the condition.

NOTE: You can only use 'euclid_apply  <rule> <args> as <X>' if the rule produces an
    existential. You should not name any propsotions introduced using 'euclid_apply' e,g,
    'euclid_apply <rule> <args> as H1'.
NOTE: It is very important that *all* non-propositional (i.e., universally quantified)
    arguments are provided to the rule when invoking 'euclid_apply'.
*TACTIC: euclid_finish *
    Attempts to resolve the proof goal using the current proof context using an SMT solver.

* euclid_assert <P> *
    Attempts to prove proposition <P> from the current proof context using an SMT solver.
        Equivalent to "have : <P> := by euclid_finish"

If you are proving an existentially quantified proposition, you can use the standard Lean tactic ' use <X>' to provide the witness <X> for the quantifier. DO NOT use the tactic 'use' if you are not proving an existentially quantified proposition.

Here is several additional tips with examples:

1. You can use standard Lean tactics such as <by_cases>, <cases>, <split_ands> and <constructor> <by_contra> to structure your proof. Specifically, you are encouraged to use "have hX: P := by" to divide the whole problems to small proposition. However, you should not use imperative Lean tactics, such as 'rw' or 'simp'. You should only use the above declarative tactics.

2. You should be careful to check the degenerate case and special cases. For example, sometimes you want to get the intersection of two lines. You may use"euclid_apply intersection_lines L1 L2 as O" but before that you should guarantee that the SMT can deduce that L1 and L2 intersects.

3. You must ensure that every step in your proof is rigorous, not only in natural language, but in LeanEuclid. For example, in the following proof, 
<error_example1>
theorem altitude_hypotenuse_similar:
  ∀ (A B C D: Point) (BC : Line),
    RightTriangle A B C ∧
    distinctPointsOnLine B C BC ∧
    foot A D BC
    → SimilarTriangles D B A A B C := by
    euclid_intros
    have h_tri_DBA : Triangle D B A := by
      euclid_finish"
    ...
<correction1>
Here if you want to claim triangle D B A, you must either prove that D is not equal to B and A, or claim it in your premise (like adding between A B D). Although in natural language it is trivial, but in this formal language you must PROVE it! In this example, instead, your method to prove h_tri_DBA should be:
    have h_tri_DBA : Triangle D B A := by
      have h4: between C D B := by
        have h5: $\angle$A:B:C < $\RightAngle$:= by
          euclid_apply triangle_angles_sum A B C
          euclid_finish
        have h6: $\angle$A:C:B < $\RightAngle$:= by
          euclid_apply triangle_angles_sum A B C
          euclid_finish
        euclid_apply acuteTriangle_foot_between A B C D BC
        euclid_finish
      euclid_finish.
NOTE: Using recursive "have"s to split the goal and make the proof neat.

Another example is:
<error_example2>
theorem apollonius_isoceles :
  ∀ (A B C D : Point) (BC : Line),
    IsoTriangle A B C ∧
    distinctPointsOnLine B C BC ∧
    Coll B D C ∧
    between B D C
    → |(A-B)| * |(A-B)| - |(A-D)| * |(A-D)| = |(B-D)| * |(C-D)| := by
  euclid_intros
  have h_A_not_on_BC : ¬(A.onLine BC) := by
    euclid_finish
  euclid_apply exists_foot A BC as H
  have h_midpoint_H : MidPoint B H C := by
    euclid_apply isoTriangle_three_lines_concidence_foot A B C H BC
    euclid_finish
  have h_tri_AHD : Triangle H A D := by
    euclid_finish

<correction2>
  Here h_tri_AHD is wrong. Since you cannot assume triangle H A D, because H may coincide with D. Instead your response shold be:
    by_cases H = D
    · ... 
    · have h_tri_AHD : triangle H A D := by
        -- H, D are on line BC, while A is not. So H, A, D are not collinear.
        euclid_finish
      ...

<error_example3>
theorem Numina_Geometry_1110 :
  ∀ (A B C H M K : Point) (AC: Line),
    (triangle A B C) ∧
    (between A H C) ∧
    (foot B H AC) ∧
    (distinctPointsOnLine A C AC) ∧
    (midpoint B M C) ∧
    (midpoint A K B)
    →
    ($\angle$ K:H:M = $\angle$ A:B:C)
    euclid_intros
    have h_tri_KHM: triangle K H M := by euclid_finish
    ...

3. "euclid_assert" make very few progress in the proof. Try to use less "euclid_assert X", but use more "have h: X := by ...".

4. When using the "*" symbol for multiplication, please ensure there is a space on both sides of the "*" symbol. For example, the correct expression should be "|(A-M)| * |(B-M)|" instead of "|(A-M)|*|(B-M)|"

5. Sometimes when chasing angles, especially using "coll_angles_eq" and "coll_supp_angles" you are encouraged to use "line_from points" to construct the between-line, for example, in the following theorem,
<error_example>
theorem median_is_half_side_implies_right_triangle:
  ∀ (A B C M : Point),
    Triangle A B C ∧
    MidPoint B M C ∧
    |(A-M)| = |(B-M)|
    → $\angle$ B:A:C = $\RightAngle$ := by
    have h_sum_BAC : $\angle$B:A:M + $\angle$M:A:C = $\angle$B:A:C := by
      euclid_apply coll_supp_angles A B M C
      euclid_finish

<correction>
In the example, "euclid_apply coll_supp_angles A B M C" will fail because the SMT cannot deduce A,B,M form a triangle. So how to prove this? actually you should add a line "euclid_apply line_from_points B C as BC" in your proof. Remember SMT cannot construct. So you should tell SMT there is a line BC, and SMT will automatically deduce A B M are not collinear. So your proof should be
theorem median_is_half_side_implies_right_triangle:
  ∀ (A B C M : Point),
    Triangle A B C ∧
    MidPoint B M C ∧
    |(A-M)| = |(B-M)|
    → $\angle$ B:A:C = $\RightAngle$ := by
    have h_sum_BAC : $\angle$B:A:M + $\angle$M:A:C = $\angle$B:A:C := by
      euclid_apply line_from_points B C as BC.
      euclid_apply coll_supp_angles A B M C
      euclid_finish

6. Take care of the order of parameter. For example, if you want to express "Right Triangle ABC with right angle ABC", you should use "RightTriangle B A C" (First parameter is rightangle) instead of "rightTriangle A B C". When apply lemma or writing formal statement, always check whether the order is align with definition. Also you should check the number of parameters. For example, "Coll" only contains three parameters. So donn't use "Coll A B C D" to represent A,B,C,D are collinear. Instead, use "Coll A B C ∧ Coll B C D"


7. At the beginning of your proof, you should firstly using "euclid_apply line_from_points X Y as XY" To obtain all the the line you needed in the problem, if the problem does not give these lines. This step is benificial to the later SMT steps.

8. When using "euclid_apply", do not add additional condition to it, for example, do not use "euclid_apply coll_supp_angles A E C B h_between_AEC hA". Instead, use "euclid_apply coll_supp_angles A E C B". SMT will automatically search whether the absent condition is satisfied.
--- End of Proof DSL ---

Your proofs can make use of the following abbreviation of geometry structure: 
--- Begin of Abbreviation ---

/-Relations-/

abbrev Coll (A B C : Point) : Prop :=
between A B C ∨ between B C A ∨ between C A B ∨ A = B ∨ A = C ∨ B = C

abbrev Triangle (A B C : Point) : Prop :=
¬ (Coll A B C)

...
...

abbrev RadicalAxis (Ω₁ Ω₂ : Circle) (L : Line) : Prop :=
∀ (A : Point), A.onLine L → Pow(A, Ω₁) = Pow(A, Ω₂)
--- End of Abbreviation ---

Also, I'll provide you the construction rules where you can construct lines, points and circles by these rules using "euclid_apply <theorem> as ...". Notice that these rules are not included in SMT. So you should construct lines, points in your proof by yourself.

--- Begin of Construction Rules ---

axiom intersection_lines : ∀ (L M : Line), L.intersectsLine M →
  ∃ a : Point, (a.onLine L) ∧ (a.onLine M)
utsideCircle α

...
...

axiom exists_distinct_point_outside_circle :
  ∀ (α : Circle) (b : Point),  ∃ a : Point, a ≠ b ∧ a.outsideCircle α

--- End of Construction Rules ---
Also, I'll provide you the theorem libarary. Use "euclid_apply" to use these theorems in theorem library.
--- Begin of Theorem Library ---

axiom triangle_area_foot :∀ (a b c d: Point) (BC: Line),b.onLine BC ∧ c.onLine BC ∧ (Triangle a b c) ∧ Foot a d BC → ($\triangle$a:b:c).area = |(a-d)| * |(b-c)|/2

...
...

theorem trapezoid_midsegment_parallel_base : ∀ (A B C D E F: Point) (AB BC CD DA EF: Line), formQuadrilateral A B C D AB BC CD DA ∧ (¬ AB.intersectsLine CD) ∧ distinctPointsOnLine E F EF ∧ MidPoint B E C ∧ MidPoint A F D →  (¬ EF.intersectsLine CD) := by
--- End of Theorem Library ---
All theorems in library are proved and you can apply them directedly. The following are few-shot example proof of the most commonly used theorems in library.
--- Few-shot Examples ---
Input1:
import Mathlib
import SystemE
import LeanGeo
namespace LeanGeo

theorem InscribedAngleTheorem_sameSide :
  ∀ (A B C O : Point) (AB: Line) (Ω : Circle), Triangle A B C ∧  distinctPointsOnLine A B AB ∧ (O.sameSide C AB) ∧ (A.onCircle Ω) ∧ (B.onCircle Ω) ∧ (C.onCircle Ω) ∧ (O.isCentre Ω)
    → $\angle$ A:O:B = $\angle$ A:C:B + $\angle$ A:C:B := by

Output1:
import Mathlib
import SystemE
import LeanGeo
namespace LeanGeo

...
...

Output5:
import Mathlib
import SystemE
import LeanGeo
namespace LeanGeo

theorem cyclic_supp_angles : ∀ (A B C D: Point) (AB:Line) (Ω : Circle), distinctPointsOnLine A B AB ∧ DistinctFourPoints A B C D ∧ A.onCircle Ω ∧ B.onCircle Ω ∧  C.onCircle Ω ∧ D.onCircle Ω ∧ C.opposingSides D AB → $\angle$B:C:A + $\angle$B:D:A = $\RightAngle$ + $\RightAngle$ := by
  euclid_intros
  euclid_apply exists_centre Ω as O
  by_cases O.sameSide C AB
  · euclid_assert O.opposingSides D AB
    euclid_apply InscribedAngleTheorem_sameSide A B C O AB Ω
    euclid_apply InscribedAngleTheorem_opposingSides A B D O AB Ω
    euclid_finish
  · by_cases O.onLine AB
    · euclid_apply ThalesTheorem A B C O Ω
      euclid_apply ThalesTheorem A B D O Ω
      euclid_finish
    · euclid_apply InscribedAngleTheorem_sameSide A B D O AB Ω
      euclid_apply InscribedAngleTheorem_opposingSides A B C O AB Ω
      euclid_finish
-- End of Few-shot Examples ---

IMPORTANT: Your response should be started with 
"import Mathlib
import SystemE
import LeanGeo
namespace LeanGeo

theorem ..." You should restate the theorem that you want to prove in formal language, give a complete proof of the theorem.
Now, please prove the following theorem:
<formal statement>
\end{lstlisting}
\end{document}